\documentclass[letterpaper]{article} 
\usepackage{aaai2026}
\usepackage{times}  
\usepackage{helvet}  
\usepackage{courier}  
\usepackage[hyphens]{url}  
\usepackage{graphicx} 
\usepackage{natbib}  
\usepackage{caption} 
\usepackage[utf8]{inputenc}
\usepackage{newunicodechar}
\newunicodechar{，}{,}

\usepackage{algorithm}
\usepackage{algorithmic}

\usepackage{newfloat}
\usepackage{listings}
\usepackage{amsmath} 
\usepackage{bm}
\usepackage{amssymb}
\usepackage{multirow} 
\usepackage{booktabs}
\usepackage{array}
\usepackage{amsfonts}
\usepackage{xcolor}

\DeclareCaptionStyle{ruled}{labelfont=normalfont,labelsep=colon,strut=off} 
\floatstyle{ruled}
\newfloat{listing}{tb}{lst}{}
\floatname{listing}{Listing}
\title{FantasyTalking2: Timestep-Layer Adaptive Preference Optimization for Audio-Driven Portrait Animation}
\author{
    MengChao Wang\equalcontrib,
    Qiang Wang\equalcontrib,
    Fan Jiang\footnote{Project Leader},
    Mu Xu
}
\affiliations{
    AMAP, Alibaba Group\\

    {wangmengchao.wmc, yijing.wq, frank.jf, xumu.xm}@alibaba-inc.com
}

\usepackage{bibentry}

\begin{document}

\maketitle

\begin{abstract}
Recent advances in audio-driven portrait animation have demonstrated impressive capabilities. However, existing methods struggle to align with fine-grained human preferences across multiple dimensions, such as motion naturalness, lip-sync accuracy, and visual quality. This is due to the difficulty of optimizing among competing preference objectives, which often conflict with one another, and the scarcity of large-scale, high-quality datasets with multidimensional preference annotations. To address these, we first introduce Talking-Critic, a multimodal reward model that learns human-aligned reward functions to quantify how well generated videos satisfy multidimensional expectations. Leveraging this model, we curate Talking-NSQ, a large-scale multidimensional human preference dataset containing 410K preference pairs. Finally, we propose Timestep-Layer adaptive multi-expert Preference Optimization (TLPO), a novel framework for aligning diffusion-based portrait animation models with fine-grained, multidimensional preferences. TLPO decouples preferences into specialized expert modules, which are then fused across timesteps and network layers, enabling comprehensive, fine-grained enhancement across all dimensions without mutual interference. Experiments demonstrate that Talking-Critic significantly outperforms existing methods in aligning with human preference ratings. Meanwhile, TLPO achieves substantial improvements over baseline models in lip-sync accuracy, motion naturalness, and visual quality, exhibiting superior performance in both qualitative and quantitative evaluations. Ours project page: \url{https://fantasy-amap.github.io/fantasy-talking2/}
\end{abstract}

\section{Introduction}

Audio-driven portrait image animation aims to synthesize realistic human speech videos from a reference image and driving audio. Recent advances \cite{chen2025hunyuanvideo,wang2025FantasyTalking,gan2025omniavatar,lin2025omnihuman,ji2025sonic,cui2024hallo3} have achieved notable improvements in facial expressions, motion diversity, and visual quality. However, critical challenges persist in generating high-fidelity animations, including difficulties in achieving perceptually natural lip-sync, the presence of obvious artifacts in complex local features (e.g., facial attributes and hand structures), and a failure to generate human motions aligned with user preferences. In language modeling and image generation, learning from human preferences \cite{ouyang2022training,rafailov2023direct,zhou2023beyond,wallace2024diffusion} has proven highly effective for enhancing generation quality and aligning models with user expectations.

However, applying such preference-driven alignment strategies to audio-driven portrait animation faces challenges. A key obstacle is the lack of large-scale, high-quality multidimensional preference data and reward models. Current methods like Hallo4 \cite{cui2025hallo4} and AlignHuman \cite{liang2025alignhuman} primarily rely on manually annotated preference data, a costly and time-consuming process that severely limits data scale. This consequently constrains model generalization for complex motions, special pronunciations, and expression variations.

Another critical barrier is the difficulty in fine-grained preference alignment due to conflicts between multi-preference objectives. Most multidimensional preference optimization methods use linear scalarization \cite{li2020deep,liu2025videodpo,liu2025improving} to combine multidimensional rewards into a composite score, enabling reuse of standard Direct Preference Optimization (DPO) \cite{rafailov2023direct}. Yet human preferences are complex and diverse, often involving conflicting goals \cite{wu2025densedpo}: a sample with better motion may exhibit poorer lip alignment, while one with superior lip alignment may demonstrate inferior motion. This makes linearly combined rewards inadequate for addressing all preferences \cite{zhou2023beyond}. 

To address these limitations, we first introduce Talking-Critic, a multidimensional video reward model designed to learn fine-grained human preferences, enabling the construction of large-scale, multidimensional preference datasets. Building upon this model, we evaluate outputs from four state-of-the-art portrait animation methods and curate Talking-NSQ, a large-scale preference dataset containing approximately 410k annotated samples, labeled automatically using Talking-Critic. The dataset includes detailed annotations on Motion Naturalness (MN), Lip Synchronization (LS), and Visual Quality (VQ), capturing the key factors that users consider when assessing generated portrait videos.

Meanwhile, many studies \cite{liang2024step,wang2024controllable} reveal that diffusion models exhibit distinct inherent biases across denoising timesteps. The initial timesteps determine the overall motion dynamics and structure, whereas the later timesteps are responsible for refining fidelity and fine-grained details. Furthermore, different layers of image and video diffusion models contribute to different dimensions of the generated results \cite{avrahami2025stable,chen2024towards}. Some critical layers significantly impact content generation, while others affect clarity and detail representation. These observations suggest that both network layers and denoising timesteps are intrinsically linked to human preferences, highlighting the necessity for layer-wise and timestep-wise preference modeling for diffusion models. 

\begin{figure*}[t]
  \includegraphics[width=\linewidth]{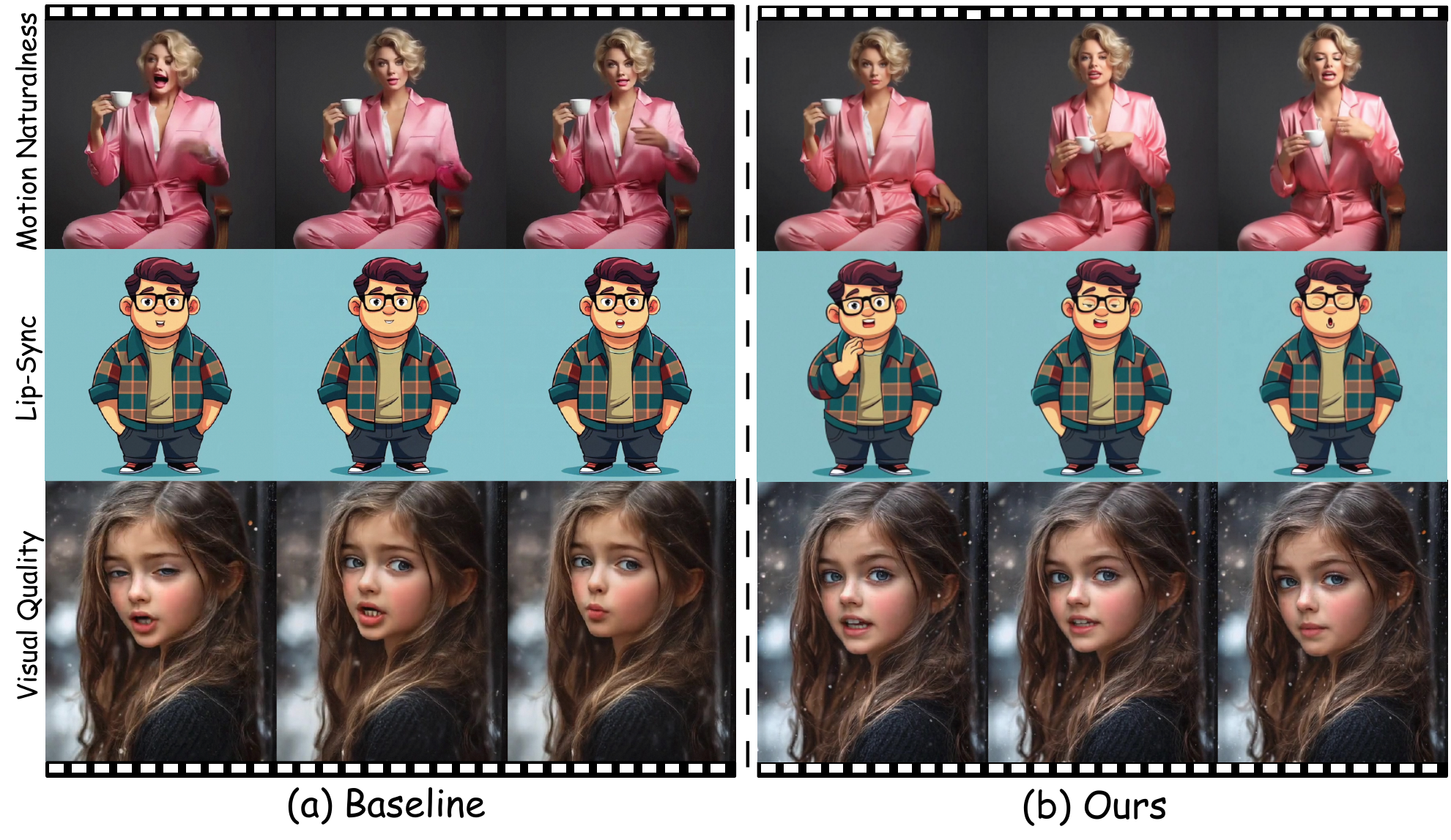}
  \caption{Comparison of the baseline and our method.}
  \label{fig:fig0}
\end{figure*}

Therefore, we propose a dual-stage Timestep-Layer adaptive multi-expert Preference Optimization (TLPO) for diffusion models. First, we employ a multi-expert approach by training lightweight LoRA \cite{hu2022lora} modules independently. Each module is specialized to optimize a specific dimension of the output, namely motion naturalness, lip-sync, and visual quality. Subsequently, we propose a fusion gate mechanism to dynamically adjust the weight distribution of each expert across timesteps and network layers. This achieves fine-grained multi-objective collaborative optimization, effectively resolving preference conflicts and overfitting to dominant preferences in traditional optimization, while enhancing overall expressiveness and human-likeness in audio-driven portrait animation generation. Figure \ref{fig:fig0} visually demonstrates the improvements of our method over the baseline across all evaluated dimensions. Our contributions can be summarized as follows:

\begin{itemize}
    \item We propose Talking-Critic, a unified multimodal reward model that accurately quantifies the alignment between generated portrait animations and multidimensional human expectations.
    \item We introduce Talking-NSQ, a large-scale portrait animation preference dataset containing 410K samples, which systematically aligns user preferences regarding audio-visual synchronization, visual quality, and motion naturalness.
    \item We propose a novel preference alignment method, termed TLPO, that adaptively integrates multiple preference objectives across timesteps and network layers. Extensive experiments demonstrate that our approach significantly outperforms existing baselines across multiple metrics.
\end{itemize}

\section{Related Work}

\subsection{Audio-Driven Portrait Animation}
Portrait animation is a highly active research area that utilizes driving signals such as video \cite{wang2025fantasyportrait,guo2024liveportrait}, audio \cite{chen2025echomimic, shen2023difftalk}, pose \cite{hu2024animate, zhu2024champ}, or identity information \cite{yuan2025identity, zhang2025fantasyid, liu2025phantom} to generate vivid portrait animations from static images.  Within the subfield of audio-driven portrait animation, early approaches \cite{ma2023dreamtalk,wei2024aniportrait,zhang2023sadtalker} leveraged 3D Morphable Models (3DMMs) \cite{egger20203d} as an intermediate representation to bridge audio and video. While 3DMMs effectively capture facial geometry and expression variations, their expressive power is inherently limited, constraining the modeling of subtle expressions and the achievement of high realism. Driven by the advancement of diffusion models, recent methods \cite{kong2025let,gan2025omniavatar,huang2025bind,chen2025hunyuanvideo} bypass intermediate representations, directly synthesizing high-quality temporal sequences from audio and a static image. These methods typically build upon large-scale, pre-trained video generative models \cite{wan2025wan,kong2024hunyuanvideo}. By incorporating audio conditioning, they generate videos exhibiting audio-visual synchronization. However, although these methods successfully integrate the audio signal, the domain gap between the data used for pre-training and subsequent fine-tuning can compromise the generative quality of the base model to some extent. Our work aims to enhance audio-driven portrait animation frameworks through reinforcement learning, utilizes a carefully designed critic model and preference data to further improve audio-visual alignment, visual quality, and the naturalness of motion in the generated sequences. 

\subsection{Human Preference Alignment}
Human preference alignment, which aims to align model outputs with human preferences, has proven effective in both language \cite{dubey2024llama,mehta2024openelm,stiennon2020learning} and vision models \cite{wallace2024diffusion,xu2023imagereward,liu2025improving}. Among alignment methods, Direct Preference Optimization (DPO) \cite{rafailov2023direct} is widely adopted \cite{liu2025videodpo,liu2024mia,wallace2024diffusion}. DPO optimizes the log probability ratio between preferred and non-preferred responses while constraining output deviations from the original distribution using a reference model. In the context of audio-driven video generation, both Hallo4 \cite{cui2025hallo4} and AlignHuman \cite{liang2025alignhuman} employ preference optimization techniques to enhance portrait animation. However, these approaches rely heavily on large-scale, manually annotated preference data pairs and do not incorporate or optimize a critic model. Consequently, their ability to satisfy complex and diverse user requirements remains limited. Our method construct a critic model to evaluate preference data pairs, eliminating the need for extensive manual annotation. Furthermore, we introduce a dual-perception strategy across timesteps and network layers to more effectively integrate multiple preference objectives.

\begin{figure*}[h]
  \includegraphics[width=\linewidth]{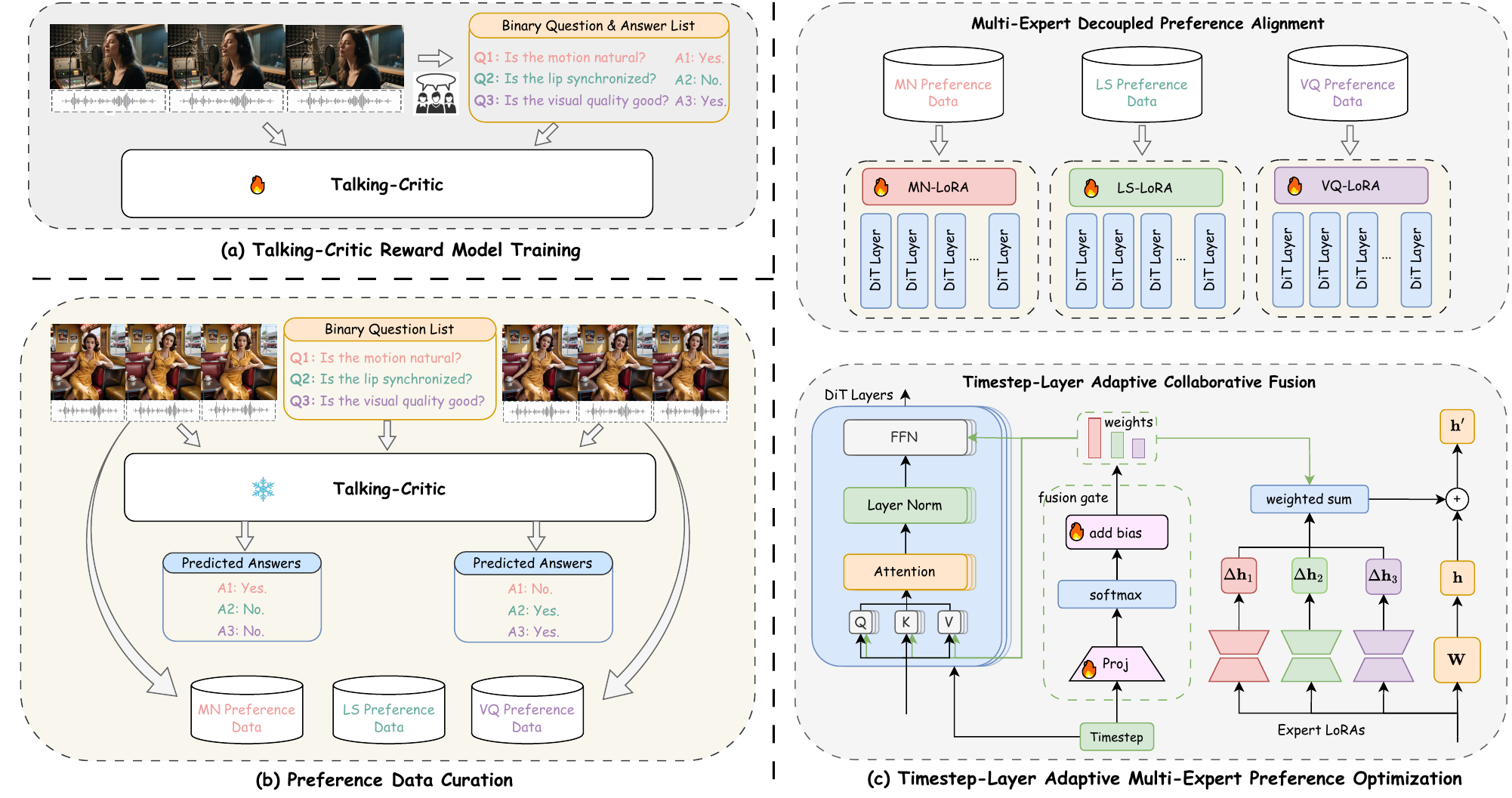}
  \caption{\textbf{The Overview of FantasyTalking2.}}
  \label{fig:overview}
\end{figure*}

\section{Method}
\subsection{Preliminary} 
\subsubsection{Base Model.}We begin by presenting our foundational model for audio-driven portrait animation, known as FantasyTalking \cite{wang2025FantasyTalking}. Built upon the pre-trained Wan2.1 \cite{wan2025wan} model, it comprises a 3D Variational Autoencoder (VAE) \cite{kingma2013auto} and a Latent Diffusion Transformer (DiT) \cite{peebles2023scalable}. The model operates in latent space rather than pixel space, utilizing the pre-trained VAE to establish bidirectional mapping between these domains. Specifically, the VAE encoder $E$ converts input video data $x$ into latent representations $z = E(x)$. The decoder $D$ reconstructs latent tokens back to video space. During training, Gaussian noise $\epsilon$ is incrementally added to $z$ via a forward process, generating noisy latent variables defined as: $z_t = tz + (1 - t){\epsilon}$, where $t \in [0, 1]$ is sampled from a logit-normal distribution. To adapt this image-to-video foundation model for audio-driven character animation, FantasyTalking utilizes Wav2Vec 2.0 \cite{baevski2020wav2vec} to extract audio features, and inject via cross-attention layers into each DiT block, enabling audio-conditioned portrait driving. 

\subsubsection{DPO for Flow Matching.} While diffusion models learn to generate human-meaningful images through reconstruction, this capability alone does not ensure alignment with human preferences and requirements. DPO provides a framework for aligning generative models with human preferences. By training on pairs of generated samples annotated with positive/negative labels, the model learns to assign higher probabilities to preferred outputs and lower probabilities to dispreferred ones. Flow-DPO \cite{liu2025improving} applies DPO to flow matching \cite{lipman2022flow}, by reformulating the DPO objective, the loss is defined as:

\begin{equation}
\mathcal{L} = - \mathbb{E}\Big[\log\sigma \Big(-\frac{\beta}{2}\big(L(x_t^w, t)-L(x_t^l, t)\big)\Big)\Big]
\label{eq:loss_flowdpo}
\end{equation}

\noindent where $\beta$ is a temperature coefficient, $L(x_t^w, t)$ and $L(x_t^l, t)$ respectively represent the losses for positive and negative parts, and $x_t^w$ and $x_t^l$ denote winning and losing samples respectively. 
This loss encourages the generation of preference-aligned samples, promoting coherent motion trajectories and producing more natural, realistic expressions.

\subsection{Talking-Critic Reward}
Previous video reward modeling approaches primarily leveraged vision-language models for training \cite{he2024videoscore,wang2024lift,xu2024visionreward}. In contrast, our audio-driven portrait animation task requires multimodal inputs encompassing text, video, and audio modalities, rendering conventional vision-language models inadequate for reward modeling. Benefiting from recent advances in unified Vision-Audio-Language Models (VALMs), significant breakthroughs have been achieved in multimodal understanding and alignment. We employ the Qwen2.5-Omni \cite{xu2025qwen2} as our foundational model, which introduces TMRoPE, this position embedding method organizes audio and video frames into chronologically interleaved structures, enabling exceptional audio-visual alignment.

As shown in Figure \ref{fig:overview}(a), to harness the full potential of Qwen2.5-Omni for evaluating portrait animations, we adapt it into a reward model through specialized instruction fine-tuning.  We construct a preference dataset encompassing three key dimensions: MN, LS, and VQ.  During its construction, we implement a rigorous balanced sampling strategy to ensure an equal number of positive and negative samples for each comparison, thereby enabling the model to learn human preferences without bias.  The resulting fine-tuned reward model provides reliable guidance signals for downstream tasks, such as DPO.

\subsection{Timestep-Layer adaptive multi-expert Preference Optimization}
For multi-objective preference optimization, existing methods \cite{zhou2023beyond, liu2025videodpo} obtain an aggregate score for each sample through various strategies, creating positive/negative pairs that reflect overall quality. This approach handles all preference objectives uniformly. However, it often leads to over-optimization in some dimensions at the expense of performance in others \cite{xu2024visionreward, liang2025alignhuman}. Specifically, In human portrait animation, a sample ranked best overall may exhibit poor lip-sync accuracy, while a sample with the lowest overall score might excel in this regard. This conflict among fine-grained preferences impedes effective, granular alignment and hinders the model's ability to learn along less prominent dimensions. To address this, we propose a two-stage training strategy. As shown in Figure~\ref{fig:overview}(c), the first stage learns decoupled preferences via a multi-expert alignment approach. Second, we introduce a timestep-layer adaptive fusion mechanism to effectively integrate these diverse preferences for robust multi-objective alignment.

\subsubsection{Multi-Expert Decoupled Preference Alignment.} To address preference competition conflicts, we first perform independent preference alignment through specialized experts, which involves three lightweight expert LoRA modules.
The first is the Motion Naturalness Expert $E_m$, which aims to ensure fluid and natural body movements. The second is the Lip Synchronization Expert $E_l$, dedicated to optimizing the coordination between audio and visual cues. The third is the Visual Quality Expert $E_v$, designed to improve the fidelity of individual frames.

Each expert module integrates into all linear layers of every DiT block. Since each specializes in a single dominant preference dimension, they achieve efficient convergence.

Noting lip synchronization exclusively concerns the mouth region, to reduce the difficulty of preference alignment and prevent introducing extraneous irrelevant preferences, we utilize MediaPipe \cite{lugaresi2019mediapipe} to extract precise lip masks in pixel space, then project them to latent space through trilinear interpolation, forming our lip-focused constraint mask $M$. Consequently, the training loss $\mathcal{L}$ in Eq. \ref{eq:loss_flowdpo} for the lip-sync expert is reweighted as:
\begin{equation}
    \mathcal{L}_c = M \odot \mathcal{L}
    \label{eq:loss2}
\end{equation}

For the motion naturalness expert LoRA and the visual quality LoRA, we perform the preference loss $\mathcal{L}$ across all pixel domains. In the end, we obtained three expert modules.

\subsubsection{Timestep-Layer Adaptive Collaborative Fusion.} Since each expert undergoes isolated dimension-wise optimization with segregated data, naively integrating them for inference may cause conflicting preferences among experts, degrading overall performance. 
Prior research has established that generative preferences differ across denoising timesteps \cite{liang2024step, wang2024controllable} and that DiT layers serve distinct functional roles \cite{avrahami2025stable, chen2024towards}. These findings motivate our design of a timestep-layer adaptive fusion strategy, enabling collaborative alignment of multi-expert modules.

Specifically, we employ a timestep-layer adaptive fusion gate to dynamically tune LoRA preference weights across DiT layers and timesteps. We integrate a lightweight, parameter-efficient fusion gate into all linear layers of DiT blocks. This gate modulates the influence of each LoRA module using the current denoising timestep $t$. As shown in Figure \ref{fig:overview}(c), for the $l$-th layer, the fusion gate takes the timestep embedding $t_\text{emb}$ and projects it to fusion weights:
\begin{equation}
    \bm{w}^{l} = \text{softmax}(W_\text{gate}^{l} t_\text{emb}) + \bm{b}^{l}
\end{equation}
\noindent where $ W_{\text{gate}}^{l} \in \mathbb{R}^{k \times d}$, $t_\text{emb} \in \mathbb{R}^{d \times 1} $ and $\bm{b}^l \in \mathbb{R}^{k \times 1}$. $k$ is the number of expert LoRA modules. In our implementation, $k=3$. $ W_{\text{gate}}$ and $\bm{b}^l$ are both learnable parameters. Crucially, since $k \ll d$ and $k \ll r$ (where $r$ is the rank of LoRA), the fusion gate introduces only negligible parameters compared to the LoRA modules themselves.

Once the layer-level and timestep-wise weight vector $\bm{w}^l$ is produced, it is broadcast to every linear sub-layer inside DiT block $l$ that carries LoRA adapters. The fusion of activations for such a layer is then performed as:
\begin{equation}
\bm{h}' = \bm{h} + \Delta \bm{h} \bm{w}^{l},
\end{equation}
\noindent where $\bm{h}$ represents the output of the $l$-th layer in the frozen DiT backbone, $\Delta \bm{h}$ denotes the delta from each expert LoRA at the same layer. During fusion training, we utilize full-dimension preference pairs, i.e., samples where the positive example is superior to its negative counterpart along all considered dimensions. During inference, the fusion gate dynamically adjusts weights $\{\bm{w}_i^{l}\}_{i=1}^k$ for per layer $l$ and timestep $t$, enabling adaptive coordination of specialized LoRAs throughout the denoising process. 

This timestep-layer dynamic fusion continuously rebalances expert contributions, resolving conflicts and preventing single-metric dominance. By promoting collaboration over competition, it drives the model toward Pareto-optimal outputs \cite{deb2011multi}.

\begin{table*}[h]
    \centering
    \small
    \begin{tabular}{>{\centering\arraybackslash}p{2.35cm}|
                    >{\centering\arraybackslash}p{1.10cm}|
                    >{\centering\arraybackslash}p{1.10cm}|
                    >{\centering\arraybackslash}p{1.10cm}|
                    >{\centering\arraybackslash}p{1.30cm}|
                    >{\centering\arraybackslash}p{1.10cm}|
                    >{\centering\arraybackslash}p{1.10cm}|
                    >{\centering\arraybackslash}p{1.10cm}|
                    >{\centering\arraybackslash}p{1.10cm}}

        \toprule
        \textbf{Method} & \textbf{HKC}\ $\uparrow$ & \textbf{HKV}\ $\uparrow$ & \textbf{SD}\ $\uparrow$ & \textbf{Sync-C}\ $\uparrow$ & \textbf{FID}\ $\downarrow$ & \textbf{FVD}\ $\downarrow$ & \textbf{IQA}\ $\uparrow$ & \textbf{AES}\ $\uparrow$  \\
        
        \midrule 
FantasyTalking & 0.838 & 30.142 & 13.783 & 3.154 & 43.137 & 483.108  & 3.685 &  2.980  \\
HunyuanAvatar & 0.883 & 37.336 & 14.812 & 4.370 & 40.063 & 475.770  & 3.758 &  2.953 \\
OmniAvatar  & 0.845 & 30.058 & 13.860  & 5.452 & 36.604 & 394.099 & 3.929 &  3.109\\
MultiTalk & 0.857 & 40.371 & 14.683 & 5.668 & 37.839 & 362.591  & 4.027 &  3.184\\
        \midrule 
Ours & \textbf{0.895} & \textbf{41.924} & \textbf{15.188} & \textbf{5.704} & \textbf{35.438} & \textbf{341.181} & \textbf{4.071} & \textbf{3.236}\\

        \bottomrule
    \end{tabular}
    \caption{\textbf{Quantitative comparisons with baselines.}}
    \label{tab:quantitative}
\end{table*}

\section{Experiments}
\subsection{Dataset Construction} \label{sec:data_construct}

\subsubsection{Multidimensional Reward Data Collection.} To train our Talking-Critic reward model, we constructed a high-quality, multi-dimensional human preference dataset. This dataset comprises both real and synthetic data, with binary preference annotations provided by professional annotators across MN, LS, and VQ dimensions. Specifically, we sourced approximately 4K real-world video clips from OpenHumanVid \cite{li2025openhumanvid}. To maximize sample diversity, we also generated 6K synthetic videos using four state-of-the-art (SOTA) audio-driven portrait models \cite{chen2025hunyuanvideo, kong2025let, gan2025omniavatar, wang2025FantasyTalking} with random classifier-free guidance scales \cite{ho2022classifier}. Subsequently, all videos were evaluated by human annotators based on dimension-specific, binary-choice questions. Each sample was independently assessed by three annotators. In cases of disagreement, a fourth senior annotator was consulted to arbitrate and make the final decision. This meticulous process yielded a multi-dimensional preference dataset of approximately 10K samples. Furthermore, we created a validation set of 1K samples following the identical procedure.
\label{preference collection}
\subsubsection{Preference Data Collection.} 
As shown in Figure~\ref{fig:overview}(b), we propose a fully automated pipeline to construct a large-scale, multi-dimensional preference dataset \textbf{Talking-NSQ} for multi-expert preference training, culminating in 410K annotated preference pairs. Specifically, for each input audio clip and reference image, we generate candidate videos using the same set of SOTA models. Each model produces four video variants per input to ensure diversity. We then employ our pre-trained Talking-Critic to score these videos across the three distinct dimensions and construct corresponding positive-negative pairs. This dimensional decoupling allows a single video to contribute to multiple preference sets, significantly enhancing data utilization efficiency. This process yielded 180K pairs for motion naturalness, 100K for lip-sync accuracy, and 130K for visual quality.

Furthermore, for the timestep-layer adaptive fusion training stage, we constructed 18K full-dimension preference pairs.   This was achieved by introducing controlled degradations to high-quality real videos.  We randomly select the four SOTA models to synthesize new videos based on real videos.   Then, we create preference pairs by matching the original high-quality real video as the positive sample with the newly generated, degraded video as the negative sample.

\subsection{Reward Learning}
\subsubsection{Training Setting.} We utilize Qwen2.5-Omni \cite{xu2025qwen2} as the backbone for our reward model, conducting supervised fine-tuning using the multidimensional reward data collected in Sec. \ref{sec:data_construct}. To adapt the model, LoRA \cite{hu2022lora} is applied to update all linear layers within Qwen2.5-Omni Thinker while keeping visual and audio encoder parameters fully frozen. The training process employs a batch size of 32 with a learning rate of $2\times10^{-6}$ over three epochs, requiring approximately 48 A100 GPU hours.

\subsubsection{Evaluation Protocols and Baseline.} 
We evaluate preference alignment accuracy of Talking-Critic using our curated 1K human-annotated test set, with comparisons against the baseline Qwen2.5-Omni model. We further employ Sync-C \cite{chung2016out} for lip-sync accuracy assessment, visual quality (IQA) score \cite{wu2023q} for visual quality evaluation, and employ SAM \cite{kirillov2023segment} to segment foreground characters from frames while separately measuring optical flow scores \cite{teed2020raft} to evaluate Subject Dynamics (SD) for character motion comparison. For Sync-C, aesthetics, and SD metrics, optimal decision thresholds are automatically determined by maximizing accuracy in distinguishing high-quality from low-quality samples.

\begin{table}[h]
    \centering
    \small
    \begin{tabular}{>{\centering\arraybackslash}p{1.55cm}|
                    >{\centering\arraybackslash}p{1.64cm}|
                    >{\centering\arraybackslash}p{1.50cm}|
                    >{\centering\arraybackslash}p{1.60cm}}

        \toprule
        \textbf{Method} & \textbf{MN Acc(\%)} & \textbf{LS Acc(\%)} & \textbf{VQ Acc(\%)}  \\
        \midrule 
SD & 78.42 & - & -   \\
Sync-C  & - & 72.34 &  -  \\
IQA & - & - & 68.85  \\
        \midrule 
Base Model & 63.15 & 52.63 & 61.24  \\

Ours & \textbf{92.50} & \textbf{86.94} & \textbf{94.67}  \\

        \bottomrule
    \end{tabular}
    \caption{\textbf{Preference accuracy on test dataset.}}
    \label{tab:reward quantitative}
\end{table}

\subsubsection{Quantitative Results.} Table \ref{tab:reward quantitative} demonstrates that our reward model achieves significantly closer alignment with human preferences across all three dimensions compared to the base model. In contrast, existing quantitative evaluation methods can only be limited to evaluation in a certain dimension and cannot precisely align with human preferences. Especially, Sync-C tends to assign higher confidence to exaggerated lip movements, whereas human annotators consistently favor natural, fluid articulation—resulting in a clear misalignment between Sync-C scores and actual human preference. 

\begin{figure*}[t]
  \includegraphics[width=\linewidth]{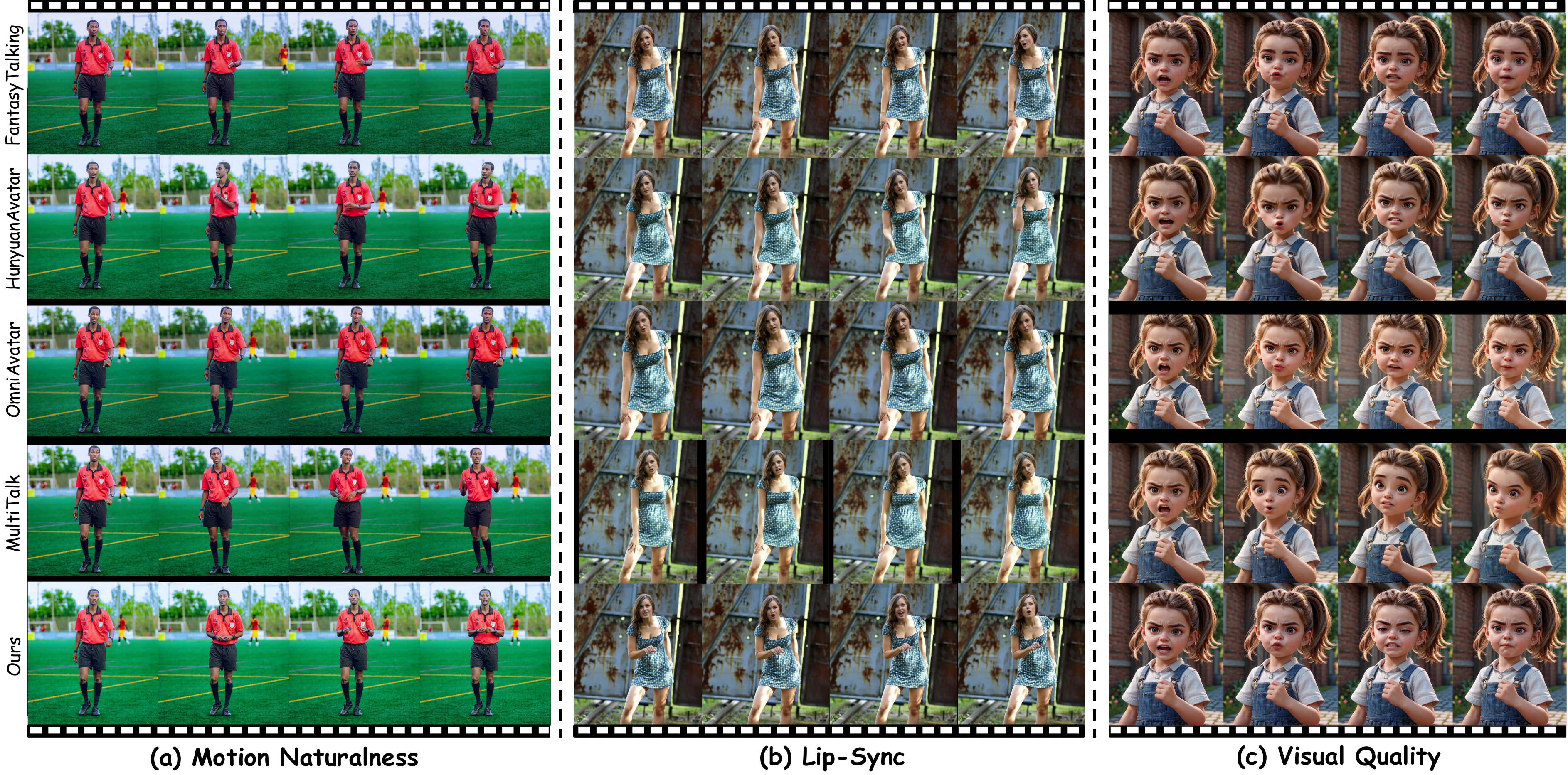}
  \caption{\textbf{Visualization results.}}
  \label{fig:fig1}
\end{figure*}

\subsection{TLPO Preference Optimization}
\subsubsection{Training Setting.} We employ the DiT-based FantasyTalking \cite{wang2025FantasyTalking} as the backbone. All training is conducted on 16 A100 GPUs optimized via AdamW. All expert modules are optimized while keeping the backbone model frozen. In the first stage of TLPO, we train  each expert LoRA module using single-dimension preference pairs, with a rank of 128. We set the learning rate to $10^{-5}$ and the $\beta$ to 5000. The MN and VQ experts undergo 10 training epochs, while the LS expert trains for 20 epochs given its complexity. In the second stage of timestep-layer adaptive multi-expert fusion, we freeze all expert LoRA layers and train minimal-parameter fusion gates using full-dimension preference pairs, with learning rate $10^{-6}$ and DPO $\beta =1000$ to balance holistic preference alignment over 5 epochs. 

\subsubsection{Evaluation Protocols and Baseline.} 
Evaluation is conducted on a benchmark test set following prior work \cite{wang2025FantasyTalking}, which cover a wide range of scenes, initial poses, and audio content.
For motion naturalness, we assess hand quality and motion richness with HKC and HKV \cite{lin2025cyberhost} , and quantify overall subject dynamics via the SD metric. 
Sync-C is used to measure the confidence of lip-sync.
For visual quality, we adopt FID \cite{heusel2017gans} and FVD \cite{unterthiner2019fvd} to assess overall generation quality and deploy q-align \cite{wu2023q} to obtain fine-grained scores on visual quality (IQA) and aesthetics (AES).
While the above metrics provide only a coarse proxy for motion naturalness, lip-sync, and visual quality, we conduct a user study for a more precise alignment check with human preference. 
We compare with the latest public state-of-the-art methods, including FantasyTalking \cite{wang2025FantasyTalking}, HunyuanAvatar \cite{chen2025hunyuanvideo}, OmniAvatar \cite{gan2025omniavatar} and MultiTalk \cite{kong2025let}, using empty prompts during inference for fair comparison.


\subsubsection{Quantitative Results.} 
Table \ref{tab:quantitative} shows our method achieves state-of-the-art results across all metrics, generating outputs with more natural motion variations, significantly improved lip synchronization, and superior overall video quality. This performance stems from TLPO preference optimization mechanism, which enables superior understanding of fine-grained human preferences in portrait animation while dynamically determining the scope and weighting of preferences according to video model denoising requirements and DiT layer characteristics. This framework achieves precise alignment with the video model's preference outputs, consequently better satisfying practical application scenarios where comprehensive quality matters.

\begin{table*}[t]
    \centering
    \small
    \begin{tabular}{>{\centering\arraybackslash}p{3.35cm}|
                    >{\centering\arraybackslash}p{1.10cm}|
                    >{\centering\arraybackslash}p{1.10cm}|
                    >{\centering\arraybackslash}p{1.10cm}|
                    >{\centering\arraybackslash}p{1.30cm}|
                    >{\centering\arraybackslash}p{1.10cm}|
                    >{\centering\arraybackslash}p{1.10cm}|
                    >{\centering\arraybackslash}p{1.10cm}|
                    >{\centering\arraybackslash}p{1.10cm}}

        \toprule
        \textbf{Method} & \textbf{HKC}\ $\uparrow$ & \textbf{HKV}\ $\uparrow$ & \textbf{SD}\ $\uparrow$ & \textbf{Sync-C}\ $\uparrow$ & \textbf{FID}\ $\downarrow$ & \textbf{FVD}\ $\downarrow$ & \textbf{IQA}\ $\uparrow$ & \textbf{AES}\ $\uparrow$  \\
        
\midrule 
w/o Timestep-wise Gating  & 0.857 & 38.240 &  14.733 & 5.509 & 39.733 &  357.963  & 3.994 &  3.122 \\
Expert-level Fusion  & 0.879 & 37.183 & 14.871 & 5.565 & 37.430 & 353.243  & 4.029 &  3.150  \\
Module-level Fusion & 0.866 & 40.968 &  15.133 & 5.649 & 36.401 &  349.053  & 4.064 &  3.198 \\
\midrule 
DPO  & 0.848 & 36.497 & 14.738  & 4.381 & 38.285 & 350.110  & 3.970 &  3.114 \\
IPO        & 0.852 & 35.118  & 14.848  & 4.375 & 37.347 & 351.415  & 4.016 &  3.163 \\
SimPO      & 0.864 & 37.935  & 14.937  & 4.546 & 37.010 & 350.935  & 4.054 &  3.217 \\
\midrule 
LoRA Rank 32 & 0.886 & 40.765 & 15.163 & 5.546 & 36.462 & 346.629  & 4.039 &  3.149\\
LoRA Rank 64 & 0.890 & 40.824 & 15.169 & 5.695 & 35.469 & 343.494  & 4.053 &  3.225 \\
LoRA Rank 256 & 0.893 & \textbf{41.930} & 15.183 & \textbf{5.704} & 35.440 & \textbf{341.174}  & 4.069 &  3.232 \\
\midrule 
Baseline & 0.838 & 30.142 & 13.783 & 3.154 & 43.137 & 483.108  & 3.685 &  2.980  \\
Ours & \textbf{0.895} & 41.924 & \textbf{15.188} & \textbf{5.704} & \textbf{35.438} & 341.181 & \textbf{4.071} & \textbf{3.236}\\
        \bottomrule
    \end{tabular}
    \caption{\textbf{Quantitative results of ablation on key designs.}}
    \label{tab:ablation}
\end{table*}

\subsubsection{Qualitative Results.}
Figure \ref{fig:fig1} demonstrates comparative results across all methods. 
On the left, our TLPO model generates natural and dynamic full-body motions, while competing methods either produce static poses or exhibit exaggerated and distorted limb movements.
The middle section highlights TLPO's robust lip-sync performance even in challenging long-range shots, where baselines exhibit severe desynchronization and misalignment.
On the right, visual quality comparisons reveal rendering flaws in other methods. 
FantasyTalking produces noticeable artifacts, OmniAvatar suffers from overexposure and blurred details, and both HunyuanAvatar and MultiTalk lose significant facial detail. In contrast, TLPO preserves high visual fidelity and structural integrity, especially in complex facial regions.

\subsubsection{User Studies.}
To further verify the alignment of the method we proposed with human preferences, twenty-four participants were asked to rate each generated video on a 0–10 scale across the three dimensions: MN, LS, and VQ. As shown in Table \ref{tab:userstudy}, our method achieves superior ratings compared to baselines, with relative improvements of 12.7\% in lip synchronization, 15.0\% in motion naturalness and 13.7\% in visual quality over the strongest baseline (MultiTalk). This comprehensive evaluation highlights the superiority of our method in generating realistic and diverse human animations that align with human preferences.

\begin{table}[h]
    \centering
    \small
    \begin{tabular}{>{\centering\arraybackslash}p{2.35cm}|
                    >{\centering\arraybackslash}p{0.78cm}|
                    >{\centering\arraybackslash}p{0.78cm}|
                    >{\centering\arraybackslash}p{0.78cm}}

        \toprule
        \textbf{Method} & \textbf{MN} & \textbf{LS} & \textbf{VQ}  \\
        \midrule 
FantasyTalking & 5.82 & 6.81 & 6.71  \\
HunyuanAvatar & 6.78 & 6.29 & 6.25   \\
OmniAvatar  & 5.95 & 7.06 & 8.09 \\
MultiTalk & 6.81 & 7.14 & 7.40  \\
        \midrule 
Ours & \textbf{8.14} & \textbf{7.96} & \textbf{8.42}  \\

        \bottomrule
    \end{tabular}
    \caption{\textbf{User Study Results.}}
    \label{tab:userstudy}
\end{table}

\subsection{Ablation Study} \label{sec:ablation}
We explore the contribution of each proposed design through several ablation studies. First, to assess our fusion mechanism, we test a variant \textbf{without timestep-wise gating}, relying only on layer-wise fusion. We also compare our proposed fusion granularity against two alternatives: \textbf{expert-level fusion}, which assigns one weight per expert, and \textbf{module-level fusion}, which assigns weights to individual linear layers (e.g., query projections). Furthermore, we establish a native \textbf{DPO} baseline by training a single LoRA on full-dimension preference pairs. We also substitute \textbf{IPO} \cite{azar2024general} and \textbf{SimPO} \cite{meng2024simpo} to evaluate alternative preference optimization methods. Finally, we investigate the effect of the LoRA rank by varying its size in the preference modules.

\begin{figure}[h]
  \includegraphics[width=\linewidth]{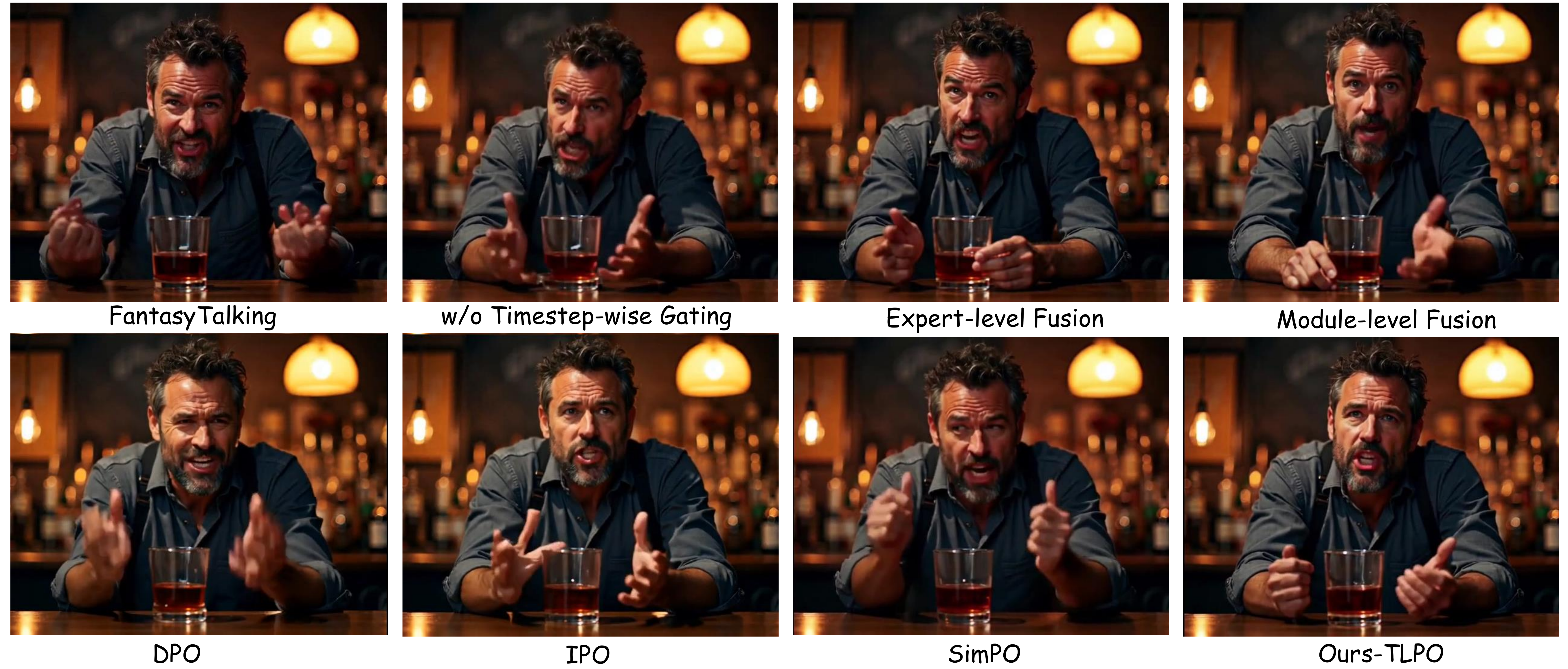}
  \caption{\textbf{Qualitative results of ablation on key designs.}}
  \label{fig:xiaorong}
\end{figure}

As shown in Table \ref{tab:ablation} and Figure \ref{fig:xiaorong}, the variant without the timestep-wise gating shows a slight improvement over the baseline but underperforms our full TLPO method. This is because different timesteps in the diffusion process have distinct optimization requirements, necessitating a flexible adjustment of the corresponding preference injection. Both expert-level and module-level fusion result in suboptimal performance. This is because different DiT layers reside in distinct manifolds and serve divergent generative functions, allowing layer-level fusion to outperform expert-level fusion. In contrast, module-level fusion introduces an excessive number of new parameters, which complicates the training process and leads to suboptimal results.

DPO and its variants achieve comparable performance with moderate visual quality improvements, yet exhibit negligible enhancement in motion naturalness and lip-sync. Although we ensured superior samples in the preference data outperform inferior ones across all dimensions, the disparity in learning difficulty between objectives introduces training ambiguity. Consequently, models prioritize optimizing the more accessible fidelity objective to mitigate synthetic artifacts, while struggling to capture nuanced motion naturalness and lip-sync preferences, resulting in limited improvements. This validates the necessity of decoupling optimization for visual quality, lip synchronization, and motion naturalness due to their inherent competing objectives.
The performance improves monotonically with the increase of LoRA rank and reaches saturation at about 128.

\section{Conclusion}
In this work, we address the challenge of balancing motion naturalness, visual fidelity, and lip synchronization in audio-driven human animation through TLPO – a novel multi-objective preference optimization framework for diffusion models. Our solution decouples competing preferences into specialized expert modules for precise single-dimension alignment, while a timestep-and-layer dual-aware fusion mechanism dynamically adapts knowledge injection throughout the denoising process. This effectively resolves multi-preference competition, enabling simultaneous optimization of all objectives without trade-offs to achieve comprehensive alignment. Qualitative and quantitative experiments demonstrate that FantasyTalking2 surpasses existing SOTA methods across key metrics: character motion naturalness, lip-sync accuracy, and visual quality. Our work establishes the critical importance of granular preference fusion in diffusion-based models and delivers a robust solution for highly expressive and photorealistic human animation.

\bibliography{aaai2026}

\end{document}